%% file: PaperForReview.tex
\documentclass[10pt,twocolumn,letterpaper]{article}

\usepackage{cvpr}              
\usepackage{appendix}
\usepackage{graphicx}
\usepackage{amsmath}
\usepackage{amssymb}
\usepackage{booktabs}
\usepackage{bbm}
\usepackage[dvipsnames]{xcolor}
\usepackage[pagebackref=true,breaklinks=true,colorlinks,bookmarks=false,citecolor=green,linkcolor=red]{hyperref}
\usepackage[accsupp]{axessibility} %
\usepackage[capitalize]{cleveref}
\crefname{section}{Sec.}{Secs.}
\Crefname{section}{Section}{Sections}
\Crefname{table}{Table}{Tables}
\crefname{table}{Tab.}{Tabs.}

\def\x{$\times$}

\usepackage{graphicx}

\usepackage{tikz}
\usepackage{comment}
\usepackage{amsmath,amssymb} 
\usepackage{color}
\usepackage[accsupp]{axessibility} 

\usepackage{tablefootnote}

\definecolor{ForestGreen}{RGB}{34,139,34}
\definecolor{OceanBlue}{RGB}{51,102,204}

\usepackage{float}
\usepackage{wrapfig}
\usepackage{textcomp, gensymb}
\usepackage{subcaption}
\usepackage{bm}
\usepackage{multirow}
\usepackage{microtype}
\usepackage{mathtools}
\usepackage{gensymb}
\usepackage{bbm}
\usepackage{chngcntr}
\usepackage{enumitem}
\usepackage{placeins}
\usepackage{etoolbox}
\usepackage{tabularx,colortbl}
\usepackage[labelfont=bf]{caption}
\usepackage{booktabs}
\usepackage{pifont}
\usepackage{algorithm}
\usepackage{listings}

\usepackage{etoolbox}
\input{cmds}

\usepackage{tikz,pgfplots}
\pgfplotsset{compat = 1.3,
	legend style={font=\scriptsize},
	legend cell align={left},
	legend style={cells={align=left}, draw=black!20},
	grid=both,
	grid style={dotted},
	tick style={draw=none},
	enlarge x limits=false,
	enlarge y limits=false,
	axis line style={draw=black!100},
	axis lines=left,
}

\pgfplotscreateplotcyclelist{mycolormarklist}{%
	chameleon3,mark=diamond*,mark options={solid, fill=chameleon3}\\
	skyblue2,mark=pentagon*,mark options={solid}\\
	scarletred3,mark=triangle*,mark options={solid}\\
	plum1,mark=square*,mark options={solid}\\
	aluminium5,mark=otimes*,mark options={solid}\\
	orange1,mark=square*,mark options={solid}\\
	skyblue3\\
	chocolate2\\
	maroon\\
	butter3\\
}
\pgfplotscreateplotcyclelist{mystylelist}{%
	densely dashdotted\\
	solid\\
	dotted\\
	densely dashed\\
	dashed\\
	dashdotted\\
	densely dotted\\
	densely dash dot dot\\
	dotted\\
	dash dot dot\\
}

\usepackage{epstopdf}
\usepackage{algorithm}
\usepackage{algorithmicx}
\usepackage{algpseudocode}
\usepackage{enumerate}

\usepackage{breqn}
\usepackage{bm}
\usepackage{t1enc}
\usepackage{colortbl}

\usepackage[USenglish,american]{babel}

\usepackage{soul}

\definecolor{carmine}{rgb}{0.59, 0.0, 0.09}

\newcommand{\app}{\raise.17ex\hbox{$\scriptstyle\sim$}}

\def\x{$\times$}
\usepackage{bbding}

\newcolumntype{x}[1]{>{\centering\arraybackslash}p{#1pt}}

\makeatletter\renewcommand\paragraph{\@startsection{paragraph}{4}{\z@}
	{.5em \@plus1ex \@minus.2ex}{-.5em}{\normalfont\normalsize\bfseries}}\makeatother

\definecolor{citecolor}{RGB}{34,139,34}
\definecolor{citecolor2}{HTML}{0071bc}
\definecolor{lightred}{RGB}{241,140,142}

\definecolor{fastcolor}{RGB}{100,178,100}
\definecolor{slowcolor}{RGB}{120,120,243}
\definecolor{expandcolor}{RGB}{244,157,78}
\definecolor{clipscolor}{RGB}{110,216,230}

\definecolor{predictioncolor}{RGB}{0,255,0}
\definecolor{labelcolor}{RGB}{255,0,0}

\definecolor{demphcolor}{RGB}{144,144,144}

\definecolor{xycolor}{RGB}{60, 120, 216}
\definecolor{xycolor}{HTML}{0071bc}
\newcommand{\xycolor}[1]{\textcolor{xycolor}{#1}}
\definecolor{wcolor}{RGB}{103, 78, 167}
\newcommand{\wcolor}[1]{\textcolor{wcolor}{#1}}
\definecolor{dcolor}{RGB}{166, 77,21}

\definecolor{gcolor}{RGB}{204, 102, 153}

\definecolor{tcolor}{RGB}{80, 200, 180}
\newcommand{\tcolor}[1]{\textcolor{citecolor}{#1}}
\definecolor{eicolor}{RGB}{153, 51, 102}
\newcommand{\eicolor}[1]{\textcolor{eicolor}{#1}}
\definecolor{grcolor}{RGB}{120, 120, 120}

\def\gaxy{\textcolor{xycolor}{$\bm{\gamma_{s}$}}}
\def\gat{\textcolor{tcolor}{$\bm{\gamma_t$}}}

\usepackage[inkscapeformat=png]{svg}

\def\cvprPaperID{13217} 
\def\confName{CVPR}
\def\confYear{2024}

\usepackage{soul}
\definecolor{defaultcolor}{gray}{.92}
\definecolor{defaultcolor2}{gray}{.90}
\sethlcolor{defaultcolor2}
\definecolor{verylightyellow}{rgb}{1,1,0.85} 

\begin{document}

\title{Multiscale Vision Transformers meet Bipartite Matching for efficient single-stage Action Localization}

\author{ Ioanna Ntinou$^{\dagger,1}$ \hspace{10pt} Enrique Sanchez$^{\dagger,2}$ \hspace{10pt}  Georgios Tzimiropoulos$^{1,2}$ \\
 $^1$Queen Mary University London, UK \hspace{25pt}
$^2$Samsung AI Center Cambridge, UK \\
{\small $^\dagger$Equal contribution} \hspace{20pt}\\
{\tt\footnotesize i.ntinou@qmul.ac.uk}  \hspace{5pt} {\tt\footnotesize kike.sanc@gmail.com} \hspace{5pt} {\tt\footnotesize g.tzimiropoulos@qmul.ac.uk}
}

\maketitle

\begin{abstract}

Action Localization is a challenging problem that combines detection and recognition tasks, which are often addressed separately. State-of-the-art methods rely on off-the-shelf bounding box detections pre-computed at high resolution, and propose transformer models that focus on the classification task alone. Such two-stage solutions are prohibitive for real-time deployment. On the other hand, single-stage methods target both tasks by devoting part of the network (generally the backbone) to sharing the majority of the workload, compromising performance for speed. These methods build on adding a DETR head with learnable queries that after cross- and self-attention can be sent to corresponding MLPs for detecting a person's bounding box and action. However, DETR-like architectures are challenging to train and can incur in big complexity.

In this paper, we observe that \textbf{a straight bipartite matching loss can be applied to the output tokens of a vision transformer}. This results in a backbone + MLP architecture that can do both tasks without the need of an extra encoder-decoder head and learnable queries. We show that a single MViTv2-S architecture trained with bipartite matching to perform both tasks surpasses the same MViTv2-S when trained with RoI align on pre-computed bounding boxes. With a careful design of token pooling and the proposed training pipeline, our Bipartite-Matching Vision Transformer model, \textbf{BMViT}, achieves +3 mAP on AVA2.2. w.r.t. the two-stage MViTv2-S counterpart. Code is available at \href{https://github.com/IoannaNti/BMViT}{https://github.com/IoannaNti/BMViT}

\end{abstract}
\input{intro.tex}

\input{related_work.tex}

\input{method.tex}

\input{experiments.tex}

\input{conclusion.tex}
\input{appendix.tex}

{\small
\bibliographystyle{ieee_fullname}
\bibliography{egbib}
}

\end{document}

%% file: cmds.tex
\newcommand{\MyMapTemplatePrefixc}[4]{\expandafter#1\csname#3#4\endcsname{#2{#4}}} 
\forcsvlist{\MyMapTemplatePrefixc {\def} {\mathcal}{c}} {A,B,C,D,E,F,G,H,I,J,K,L,M,N,O,P,Q,R,S,T,U,V,W,X,Y,Z}  
\newcommand{\MyMapTemplatePrefixtb}[5]{\expandafter#1\csname#4#5\endcsname{#2{#3{#5}}}} 
\forcsvlist{\MyMapTemplatePrefixtb {\def} {\tilde}{\mathbf}{t}} {A,B,C,D,E,F,G,H,I,J,K,L,M,N,O,P,Q,R,S,T,U,V,W,X,Y,Z}  
\forcsvlist{\MyMapTemplatePrefixtb {\def} {\tilde}{\mathbf}{t}} {0,1,a,b,c,d,e,f,g,h,i,j,k,l,m,n,o,p,q,r,s,u,v,w,x,y,z}  
\makeatletter
\newcommand\footnoteref[1]{\protected@xdef\@thefnmark{\ref{#1}}\@footnotemark}
\makeatother
\newcommand{\MyMapTemplateNoPrefix}[3]{\expandafter#1\csname#3\endcsname{#2{#3}}}
\forcsvlist{\MyMapTemplatePrefixc {\def} {\mathbf}{mb}} {0,1,a,b,c,d,e,f,g,h,i,j,k,l,m,n,o,p,q,r,s,u,v,w,x,y,z}  
\forcsvlist{\MyMapTemplatePrefixc {\def} {\mathbf}{mb}} {A,B,C,D,E,F,G,H,I,J,K,L,M,N,O,P,Q,R,S,T,U,V,W,X,Y,Z}

\forcsvlist{\MyMapTemplatePrefixc {\def} {\mathrm}{rm}} {0,1,a,b,c,d,e,f,g,h,i,j,k,l,m,n,o,p,q,r,s,u,v,w,x,y,z}  
\forcsvlist{\MyMapTemplatePrefixc {\def} {\mathrm}{rm}} {A,B,C,D,E,F,G,H,I,J,K,L,M,N,O,P,Q,R,S,T,U,V,W,X,Y,Z}    

\forcsvlist{\MyMapTemplatePrefixc {\def} {\mathcal}{mc}} {0,1,a,b,c,d,e,f,g,h,i,j,k,l,m,n,o,p,q,r,s,u,v,w,x,y,z}  
\forcsvlist{\MyMapTemplatePrefixc {\def} {\mathcal}{mc}} {A,B,C,D,E,F,G,H,I,J,K,L,M,N,O,P,Q,R,S,T,U,V,W,X,Y,Z}

\newcommand{\cmark}{\ding{51}}%
\newcommand{\xmark}{\ding{55}}%

\crefformat{section}{\S#2#1#3}
\crefformat{subsection}{\S#2#1#3}
\crefformat{subsubsection}{\S#2#1#3}

\definecolor{lightgray}{gray}{.92}

\newcolumntype{x}[1]{>{\centering\arraybackslash}p{#1pt}}

\newlength\savewidth\newcommand\shline{\noalign{\global\savewidth\arrayrulewidth
  \global\arrayrulewidth 1pt}\hline\noalign{\global\arrayrulewidth\savewidth}}

%% file: intro.tex
\section{Introduction} \label{sec:intro}

Action Localization is a challenging task that requires detecting a person's bounding box within a given frame and classifying their corresponding actions. This task shares similarities with object detection, with the particularity that the detected objects are always people, and the classes correspond to various actions that can sometimes co-occur. It poses the additional challenge that actions require temporal reasoning, as well as the fact that a person detector can contribute to false positives if a given person is not performing any of the target actions. The current golden benchmark of AVA 2.2.~\cite{li2020ava} has a fairly low mean Average Precision (mAP) compared to that of e.g. COCO~\cite{lin2014microsoft}.

\begin{figure*}[t]
  \centering
 \includegraphics[width=1\linewidth]{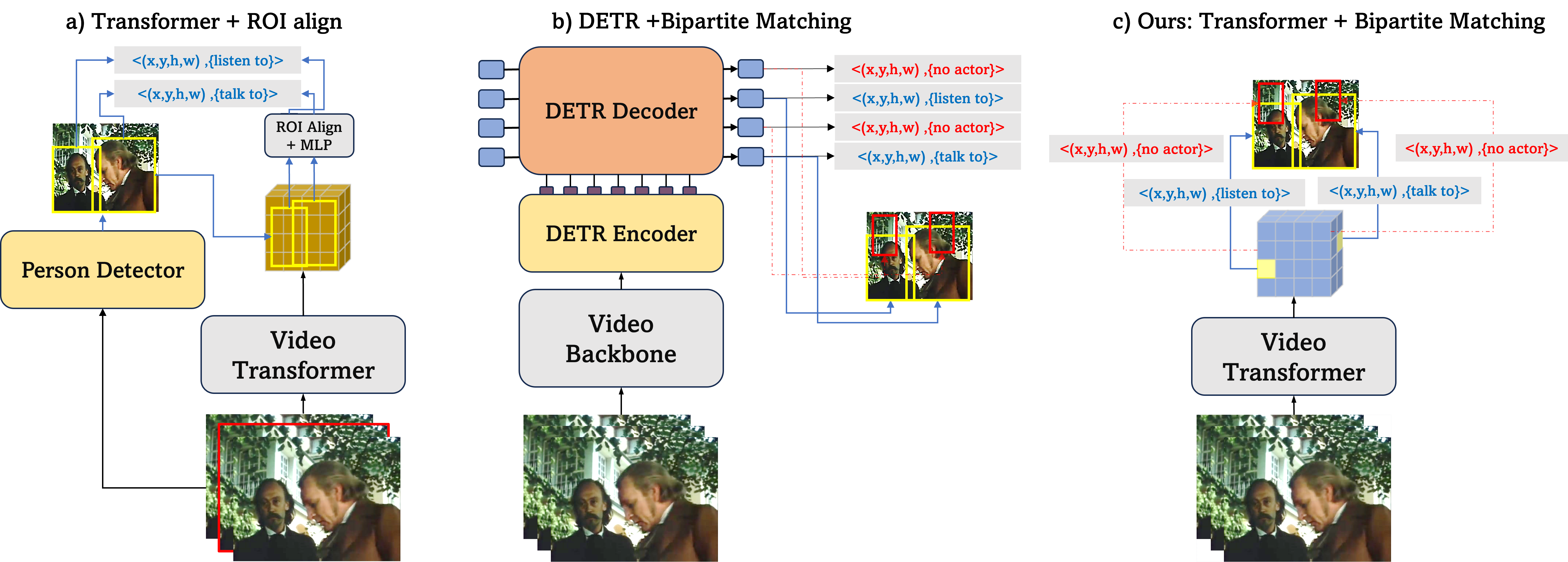}
\caption{Comparison between existing works and our proposed approach. \textbf{a)} Traditional two-stage methods work on developing strong vision transformers that are applied in the domain of Action Localization by outsourcing the bounding box detections to an external detector. ROI Align is applied to the output of the transformer using the detected bounding boxes, and the pooled features are forwarded to an MLP that returns the class predictions. \textbf{b)} Recent approaches in one-stage Action Localization leverage on the DETR capacity to model both the bounding boxes and the action classes. A video backbone produces strong spatio-temporal features that are handled by a DETR transformer encoder. A set of learnable queries are then used by a DETR transformer decoder to produce the final outputs. \textbf{c) Our method} builds a vision transformer only that is trained against a bipartite matching loss between the individual predictions given by the output spatio-temporal tokens and the ground-truth bounding boxes and classes. Our method does not need learnable queries, as well as a DETR decoder, and can combine the backbone and the DETR encoder into a single architecture. }
\label{fig:graphabs2}
\vspace{-0.2in}
\end{figure*}

Interestingly, state-of-the-art approaches achieving high absolute mAP outsource the detection task to a pre-trained Faster-RCNN~\cite{ren2015faster} and focus on large capacity networks and the use of large pre-training data (see \cref{fig:graphabs2}a)). These methods serve the purpose of boosting the absolute performance by means of accuracy but are, in many cases, prohibitive even for standard GPU accelerators. 

In this paper, we contribute to the domain of low-regime scalable models that perform both the detection and the recognition tasks. Because of its similarities with the object detection task, many recent methods targeting a single-stage model build on having a strong backbone that provides temporal features to a DETR~\cite{carion2020end} architecture~\cite{zhao2021tuber,wu2023stmixer}. A DETR architecture is an encoder-decoder transformer with learnable queries that are assigned to the ground-truth pairs of bounding box/action in a set prediction fashion (\cref{fig:graphabs2}b)).

While DETR-based architectures have shown to be efficient choices for end-to-end action localization (i.e. for joint detection and classification tasks), their design involves a video backbone and an encoder-decoder transformer architecture. This poses the question of whether there is room for further improvement in the network design. To answer this question, we draw inspiration from the recent advances in Open-World Object Detection using Vision Transformers (OWL-ViT~\cite{minderer2022simple}) and make the following \textbf{contribution}: \textit{we propose the use of a bipartite matching loss between the spatio-temporal output embeddings of a single transformer backbone and the ground-truth instances in a video clip} (\cref{fig:graphabs2}c). In this setting, the video embeddings are independent tokens that can be matched to the predictions similarly to that of DETR. This implies that a) no learnable tokens are necessary, nor a transformer decoder with self- and cross-attention, and b) the video backbone and the encoder can be merged into a single strong video transformer. Such a simple approach with a careful token selection allows us to train an MViTv2-S~\cite{li2021improved} with a simple MLP head to directly predict the bounding boxes and the action classes. Without additional elements or data, our single-stage MViTv2-S surpasses the two-stage MViTv2-S of \cite{li2021improved} that was trained using RoI align and pre-computed bounding boxes.

\noindent \textbf{Our main results} indicate that with similar FLOPs an MViTv2-S trained with bipartite matching performs better than the same MViTv2-S when trained only for action classification from precomputed bounding boxes. In addition, by simply removing the last pooling layer of MViTv2-S we obtain a +3 mAP increase on AVA2.2~\cite{li2020ava}. To the best of our knowledge, our method is the first to apply an encoder-only vision transformer for action localization with a bipartite matching loss.

%% file: related_work.tex
\section{Related Work}
\label{sec:related}

\noindent\textbf{Two-stage spatio-temporal action localization:}  Most existing works on action detection~\cite{sun2018actor,zhang2019structured,wu2019long,feichtenhofer2019,wu2020context,ulutan2020actor,pan2021actor,fan2021multiscale, wu2022, faure2022} depend on a supplementary person detector for actor localization. Typically, that is a Faster RCNN-R101-FPN~\cite{ren2015faster} detector,  originally trained for object detection on COCO~\cite{cocodataset} dataset and subsequently fine-tuned on the AVA~\cite{gu2018}, is incorporated in the action detection task. By introducing an off-the-self detector, the action detection task is simplified to an action classification problem. For actor-specific prediction, the RoIAlign~\cite{he2017} operation is applied to the generated 3D feature maps. The aforementioned standard pipeline is employed by SlowFast~\cite{feichtenhofer2019}, MViT~\cite{ li2021improved}, VideoMAE~\cite{tong2022videomae} and Hiera~\cite{ryali2023hiera} where RoI features are directly utilized for action classification. However, such features only confine information within the bounding box, neglecting any contextual information beyond it. To address the limitation, AIA~\cite{tang2020asynchronous} and ACARN~\cite{pan2021actor} employ an additional heavyweight module to capture the interaction between the actor and the context or other actors. Furthermore, to model temporal interactions, MeMViT~\cite{wu2022}, incorporated a memory mechanism on an  MViT~\cite{li2021improved} backbone. While achieving high accuracy, these methods are inefficient for real-world deployment. Our method is on par with the state-of-the-art MeMViT requiring fewer FLOPs.

\noindent\textbf{Single-stage spatio-temporal action localization:} Motivated by the aforementioned limitation of the traditional two-stage pipeline for action detection and classification, several works attempted to tackle both detection and classification in a unified framework. Some works borrow solutions developed for object detection and adapt them to action detection~\cite{zhao2021tuber, wu2023stmixer} while others simplify training through joint actor proposal and action classification networks~\cite{chen2021watch, girdhar2019video,kopuklu2019yowo, sun2018actor, Sui2022ASA}, or draw their attention to the task of Temporal Action Localization~\cite{yang2017,chao2018,shou2017,shou2016}. SE-STAD~\cite{Sui2022ASA} builds on the Faster-RCNN framework of \cite{ren2015faster} and incorporates into it the action classification task. Similarly, the Video action transformer network ~\cite{girdhar2019video} is a transformer-style action detector to aggregate the spatio-temporal context around the target actors. More recent works~\cite{zhao2021tuber,wu2023stmixer, chen2023} leverage on recent advancements of DETR in object detection, and hence, they propose to form the task using learnable queries to model both action and bounding boxes. TubeR~\cite{zhao2021tuber} proposed a DETR-based architecture where a set of queries, coined Tubelet Queries, simultaneously encode the temporal dynamics of a specific actor's bounding box as well as their corresponding actions. TubeR uses a single DETR head to model the Tubelet Queries, with a classification head that requires an extra decoder for the queries to attend again to the video features. In a similar fashion,  DETR-like fashion, STMixer~\cite{wu2023stmixer} proposes to adaptively sample discriminative features from a multi-scale spatio-temporal feature space and decode them using an adaptive scheme under the guidance of queries. EVAD~\cite{chen2023} suggests two video action detection designs: 1) Token dropout focusing on keyframe-centric spatiotemporal preservation and  2) Scene context refinement using ROI align operation and a decoder. Contrary to previous works that use a decoder or a heavy module that introduces interaction features of context or other actors, we demonstrate that a single transformer model trained directly with bipartite matching can achieve similar accuracy to more complex solutions based on DETR.

%% file: method.tex
\section{Method}
\label{sec:method}
\noindent To motivate our approach, we first depart from a standard application of Vision Transformers for Action Localization with pre-computed bounding boxes to then provide a brief description of single-stage methods that build on using bipartite matching (i.e. DETR). We then introduce our approach: a Video Transformer with bipartite matching, without learnable queries and decoder. 

\subsection{Preliminaries}
\label{ssec:problem}
\noindent The goal of Action Localization is to detect and classify a set of actions in the central frame $I_t$ of a video clip $X$ composed of $T$ frames. Because not every person in a video clip might be performing an action of interest, we distinguish between a person and an \textit{actor}, i.e. a person doing any of the $C$ target \textit{actions}. An actor at time $t$ is defined by a \textit{bounding box} ${\bf b} = [x_c, y_c, h, w]$, with $x_c,y_c$ the normalized coordinates of the box center and $h,w$ the normalized height and width of the box, and an \textit{action class} one-hot vector ${\bf a} = \{0,1\}^C$ representing the activation or not of each class. The action classes do not need to be mutually exclusive (e.g. ``talk to" and ``point to (an object)" can co-occur).

\subsubsection{Vision Transformers for Action Localization}
\label{ssec:vits}
\noindent Multi-scale Vision Transformers (MViT, ~\cite{dosovitskiy2020vit, li2021improved}) are self-attention-based architectures that operate on \textit{visual tokens} produced by dividing the input video (or image) into $\tilde{L} = T \times H \times W$ patches of size $3 \times (\tau \upsilon \nu)$, and by projecting each into $D$-dimensional embeddings through a linear or a convolutional layer. MViT architectures operate hierarchically considering many small input patches of few channels $D$, progressively increasing the patch size and the channel dimensions through pooling layers. Without loss of generality, we define a (Multi-scale ) Vision Transformer as a network $\mathcal{V}$ that produces an output set of $\tilde{L}$ tokens of $d$ dimensions from an input clip $X \in \mathbb{R}^{D \times T \times H \times W}$, as $\tilde{X} = \mathcal{V} (X) \in \mathbb{R}^{\tilde{L} \times d}$, with $\tilde{L} = t \times h \times w$, and $t \leq T$, $w \leq W$, $h \leq H$.

Generally, Vision Transformers are first pre-trained on Action Recognition datasets with video-only mutually exclusive classes, using an additional class token $X_0$ prepended to $X$. Then, Vision Transformers are adapted to Action Localization tasks by using an \textit{external} actor detector (i.e. a person detector fine-tuned to return positives on actors only) that provides a bounding box $\hat{\bf b}$. The output $\tilde{L}$ is then treated as a spatio-temporal feature map: ROI-align is done on the temporally pooled feature map $h \times w$ using $\hat{\bf b}$, and forwarded to an action classifier to produce the action class probabilities $p(\hat{\bf a})$. 

These models, known as \textit{two-stage}, compromise complexity for accuracy, resulting in solutions with prohibitive complexity. Existing works use a state-of-the-art Faster-RCNN detector to compute the bounding boxes, resulting in models with added complexity of $246$ GFLOPs, regardless of the size of the proposed architectures. While MViT has recently been proposed for object detection by adding Mask-RCNN~\cite{he2017} and Feature Pyramid Networks~\cite{lin2017feature}, the use of a single architecture to perform both detection and action classification is unexplored. Nonetheless, to our knowledge, ours is the first Vision Transformer that can do bounding box detection and action classification in a single step.

\subsubsection{DETR for Action Localization}
\label{ssec:DETR}
\noindent DETR (DEtection TRansformers~\cite{carion2020end}) formulate object detection as a bipartite matching problem where a fixed set of $L$ learnable embeddings, known as object queries, are one-to-one matched to a list of predictions of the form $\langle {\bf b},  p(\hat{{\bf a}}')\rangle$, with $\hat{{\bf a}}' = \{\hat{\bf a} , \varnothing\}$ the list of $C$ target classes (objects in this case) appended with the empty class $\varnothing$ representing a no-object class. During training, the Hungarian Algorithm~\cite{stewart2016end,Kuhn1955Hungarian} is used to map the $N < L$ objects in an image to the ``closest" predictions so that the assignments minimize a combined bounding box and class cost. The $L - N$ remaining predictions are assigned to the empty class. The learning is done by backpropagating the bounding box and class error for those outputs matched to a ground-truth object and only the class error for those assigned to the empty class (i.e. to enforce bounding boxes that do not correspond to an object to be tagged as $\varnothing$).

DETR possesses appealing properties for Action Localization, as the learnable object queries can convey object localization and action classification. The handful of proposed approaches that have ventured to apply DETR to Action Localization are faithful to the DETR configuration~\cite{zhao2021tuber,wu2023stmixer}: a 3D backbone produces video features that are fed into a transformer encoder to produce a hierarchy of features, and a transformer decoder transforms the learnable queries through self- and cross-attention layers between the queries and the encoder features, producing a fixed set of $L$ outputs. However, rather than appending the target classes with the empty class, a dedicated \textit{actor detector} head is used to determine if a bounding box corresponds to an actor or not. The outputs are now triplets of the form $\langle {\bf b}, p(\alpha), p(\hat{\bf a}) \rangle$, with $p(\alpha)$ being a two-logit vector representing the actor and the $\varnothing$ classes. The use of a DETR encoder-decoder architecture offers a good tradeoff between accuracy and complexity with the decoder alleviating part of the backbone's complexity. However, these approaches still suffer from a complex architecture and a limited number of queries that can be used to train the models.

\subsection{Our solution: Bipartite-Matching ViT}
\label{ssec:ours}
\noindent Herein, we observe that neither the backbone nor the decoder are necessary to achieve a good accuracy-speed tradeoff, and we follow the recent advances in Open-Vocabulary Object Detection(OWL-ViT~\cite{minderer2022simple}) to motivate our approach. OWL-ViT introduces a CLIP vision transformer encoder with the last pooling layer removed. Instead of pooling the output tokens to form a visual embedding to be mapped into the text embeddings as in the standard CLIP image-text matching, each of the output embeddings is forwarded to a small head consisting of a bounding box MLP and a class linear projection. This way, the $\tilde{L} = h \times w$ output tokens from the vision encoder are treated as independent output pairs $\langle \hat{\bf b},\, p(\hat{\bf a}) \rangle$. These pairs are matched to the ground truth using DETR's bipartite matching loss. In OWL-ViT, the image patches play the role of the object queries. 

To adapt OWL-ViT to the domain of Action Localization, we first note that Multi-scale Vision Transformers are a natural pool of spatio-temporal output embeddings that can be one-to-one matched to triplets $\langle \hat{\bf b}, p(\alpha), p(\hat{\bf a}) \rangle$. As we demonstrate in \cref{ssec:mvit1}, a simple MViTv2-S architecture trained with bipartite matching achieves higher accuracy than the same MViTv2-S trained for Action Localization using external bounding boxes. Notably, our approach, which we coin \textbf{Bipartite-Matching Vision Transformer}, or \textbf{BMViT}, does not add additional complexity to the backbone, given that the heads are simple MLPs. 

The output of the video transformer, as introduced in \cref{ssec:vits}, is $\tilde{X} = \mathcal{V}(X) \in \mathbb{R}^{\tilde{L} \times \tilde{D}}$, with $\tilde{L} = t \times h \times w$ corresponding to the output sequence length. For instance, MViTv2-S produces, for an input video of $16 \times 256 \times 256$, an output of $8 \times 8 \times 8$ tokens, i.e. $\tilde{L} = 512$, which outnumbers DETR-based architectures. The number of output tokens does not affect the complexity of the network as these are independently processed by three MLPs. Following OWL-ViT, we add a bias to the predicted bounding boxes to make each be centered by default on the image patch that corresponds to the 2D grid in which the output tokens would be re-arranged. As reported in \cite{minderer2022simple}, ``there is no strict correspondence between image patches and tokens representations"; however, ``biasing box predictions speeds up training and improves final performance". In our 3D space-time setting, we add the same 2D bias to all tokens on the same 2D grid along the temporal axis. 

Because each head will process the output tokens independently to produce $\hat{\bf b}$, $p(\alpha)$ and $p(\hat{\bf a})$, we can select which of these are of better use to the detection and recognition subtasks. \textit{The only technical limitation is that the outputs of each head need to be in one-to-one correspondence with those from the other heads, to form the triplets that will be matched to the ground-truth instances}. This is an important consideration because the tasks of actor detection and action classification are opposed by definition: while detecting actors requires only information from the central frame, the task of action recognition benefits from using temporal support. As we are not limited to use the same output tokens for each task, we can consider e.g. the $w \times h$ tokens corresponding to $t = \lfloor T/2 \rfloor$ to generate $\tilde{L} = w \times h$ bounding boxes $\hat{\bf b}$ and actor/no actor probabilities $p(\alpha)$, and apply temporal pooling to produce an equivalent set of $\tilde{L}$ tokens that will be used to compute the action probabilities $p(\hat{\bf a})$.  

\begin{figure}[t]
  \centering
 \includegraphics[width=1.\linewidth]{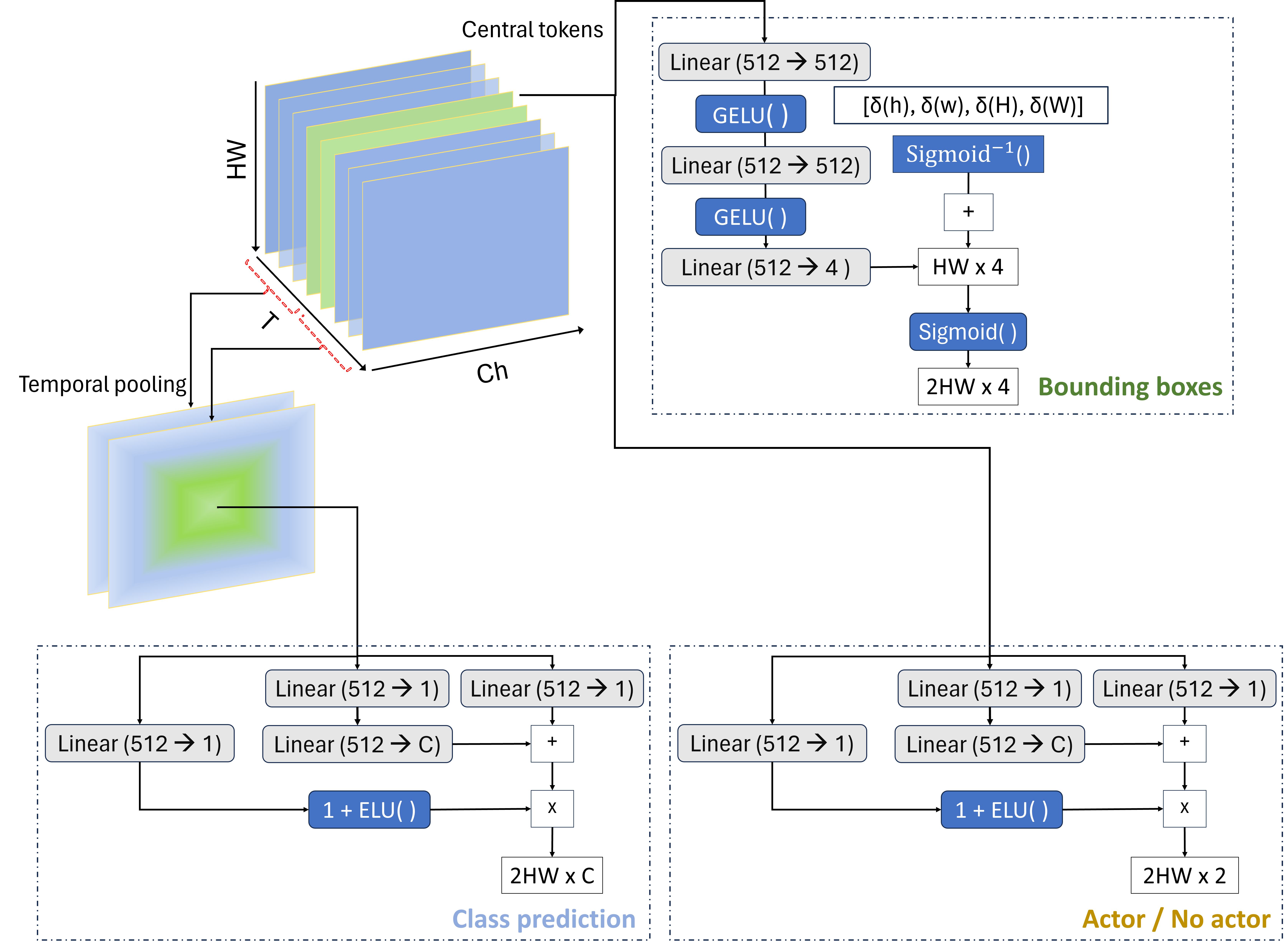}
\caption{The output spatio-temporal tokens are fed to 3 parallel heads. We use the central tokens to predict the bounding box and the actor likelihood while averaging the output tokens over the temporal axis to generate the action tokens. Each head comprises a small MLP that generates the output triplets. We depict the flow diagram for each head, following the standard OWL-ViT head~\cite{minderer2022simple}. }
\label{fig:heads}
\vspace{-0.2in}
\end{figure}

Bearing this in mind, we can handle the visual tokens in a way that benefits both tasks. First, to avoid poor detection when using MViTv2-S, we remove the last pooling layer to increase the output resolution, i.e. to produce an output of $\tilde{L} = 8 \times 16 \times 16 = 2048$ tokens. This way, we can use the $\tilde{L} = h \times w = 256$ central tokens for the detection task, and apply temporal pooling to form the equivalent $\tilde{L} = 256$ spatio-temporal tokens to be used for the classification task. We also study an alternative strategy that considers, for the bounding box and actor/no actor predictions, the tokens corresponding to both $t = \lfloor T/2 \rfloor$ \textit{and} $t = \lceil T/2 \rceil$. Concatenating both results in $\tilde{L} = 512$ tokens to perform the actor detection task. To produce the corresponding $\tilde{L} = 512$ ``action" tokens, we apply temporal pooling on the past and future tokens w.r.t. the central frame, independently. That is to say, we compute $\tilde{X}_{t<T/2}' \in \mathbb{R}^{h \times w}$ as $\tilde{X}'_{t<T/2 \, [i,j]} = (2/T) \sum_{t'<T/2} \tilde{X}_{t',i,j} \, \text{  for  } i,j \in [1...h,1...w]$, and similarly $\tilde{X}_{t>T/2}' \in \mathbb{R}^{h \times w}$. We then concatenate both to form the final set of $\tilde{L} = 512$ tokens. Note that both the ``actor" and ``action" tokens have a clear one-to-one correspondence, which is the necessary condition to produce $\tilde{L}$ triplets $\langle \tilde{\bf b}, p(\alpha), p( \tilde{\bf a}) \rangle$. As we observe in \cref{ssec:tokens}, this approach results in better performance than the one above that considers only $\tilde{L} = 256$ tokens. Notably, this strategy also outperforms that of considering the full set of $\tilde{L} = 2048$ tokens, pointing to the need of carefully choosing which tokens are more suitable for each subtask. 

\paragraph{Training} During training, the output tokens are forwarded to the corresponding heads to predict the $\tilde{L}$ triplets $\hat{y}_l = \langle \hat{\bf b}_l, p(\alpha)_l, p(\hat{\bf a})_l \rangle$ with $l \in [0,\tilde{L}-1]$. We use the Hungarian Algorithm to match these predictions to the ground-truth instances. A ground-truth instance with $N$ bounding boxes is defined as $y_j = \langle {\bf b}_j, \alpha_j=1, {\bf a}_j \rangle$ for $j < N$, and as $y_j = \langle {\bf b}_j=\varnothing, \alpha_j=\varnothing, {\bf a}_j = \varnothing \rangle$. The matching cost between a prediction $\hat{y}_i$ and a ground-truth triplet $y_j$ is defined as 
\begin{equation} 
\begin{aligned}
\label{eq:hungarian}
\mathcal{C}(\hat{y}_i, y_j) = \mathbbm{1}_{[\alpha = 1]}\mathcal{L}_{box}({\bf b}_j, \hat{\bf b}_i) - \\ \mathbbm{1}_{[\alpha = 1]} \mathcal{L}_{actor}(\alpha_j , p(\alpha_i)) - \mathbbm{1}_{[\alpha = 1]}\mathcal{L}_{class}({\bf a}_j, p({\bf a}_i)).
\end{aligned}
\end{equation}
with $\mathcal{L}_{box}$ the bounding box loss, $\mathcal{L}_{actor}$ the actor/no actor loss, and  $\mathcal{L}_{class}$ the classification loss, respectively. Following DETR~\cite{carion2020end} the bounding box loss is defined as $\mathcal{L}_{box }({\bf b}, \hat{\bf b}) = \mathcal{L}_{iou}({\bf b}, \hat{\bf b}) + \| {\bf b} - \hat{\bf b}\|_1$, with $\mathcal{L}_{iou}$ the generalized IoU loss~\cite{rezatofighi2019generalized}. The actor/no actor loss is defined as $\mathcal{L}_{actor}(\alpha_j , p(\alpha_i)) = -\alpha_j \log p(\alpha_i) + (1 - \alpha_j) \log(1 - p(\alpha_i))$, and the classification loss as:
\begin{dmath}
    \mathcal{L}_{class}({\bf a}_j, p({\bf a}_i)) = 
    -\sum_{c} {\bf a}_{j,c} \log p({\bf a}_{i,c}) + (1 - {\bf a}_{j,c}) \log (1 - p({\bf a}_{i,c})),
\end{dmath} 
where ${\bf a}_i = \{{\bf a}_{i,c} \}_{c=1...C}$ is a $C$-d vector of $1$s and $0$s representing the multi-label nature of the problem, and $p({\bf a}_{,c})$ represents the model confidence for class $c$. For $\alpha=0$, we simply set ${\bf a}_{i,c} = 0, \, \forall c$. Note that \cref{eq:hungarian} only considers the cost of the assignments w.r.t. the annotated bounding boxes (i.e. $\alpha=1$).

Once an optimal assignment $i = \sigma(j)$ $\forall i \in [0,...,\tilde{L}-1]$ is found, we compute the \textit{Hungarian Loss} as:
 
\begin{equation}
\label{eq:total_loss}
\begin{aligned}
\mathcal{L} = \sum_{j=1}^{L}[ \mathbbm{1}_{[\alpha = 1]}\left( \lambda_{iou} \mathcal{L}_{iou}({\bf b}, \hat{\bf b}) + \lambda_{L1} \| {\bf b} - \hat{\bf b}\|_1\right) - \\
\lambda_{\alpha} \mathcal{L}_{actor}(\alpha_j , p(\alpha_\sigma(j))) - \lambda_{a} \mathcal{L}_{class}({\bf a}_j, p({\bf a}_\sigma(j))) ]
\end{aligned}
\end{equation} 
where $\lambda = \{\lambda_{iou}, \lambda_{L1}, \lambda_{\alpha}, \lambda_{a}\} \in \mathbb{R}^4$ are hyperparameters.

\paragraph{Inference} During inference, we compute the $L$ output triplets, and we simply keep the detections of those for which $p(\alpha) > \theta$ with $\theta$ a hyperparameter that controls the trade-off between precision and recall.

\paragraph{Remarks} We want to highlight that our method, which resides in reformulating the training objective of vision transformers for action localization and in the observation that such approach comes with different alternatives for token selection, is amenable to different backbones, token selection design, and even output resolution. The fact that tokens are fixed and assigned to ground-truth instances also implies that we can directly resize the input frames to a fixed squared resolution, without any concern regarding losing the aspect ratio, something not possible in ROI-based approaches. This prevents our method from the need of using different views, being computationally more efficient.

%% file: experiments.tex
\input{ablation}

\section{Experimental Setup}
\label{sec:exp}
\noindent \textbf{Datasets}. We use \textbf{AVA 2.2}~\cite{gu2018, li2020ava} to conduct our main experiments and ablations, and \textbf{UCF101-24}~\cite{soomro2012ucf101} and \textbf{JHMDB51-21}~\cite{jhuang2013towards} to demonstrate the generalising capabilities of our approach. \textbf{AVA 2.2}~\cite{gu2018, li2020ava} is a long-tail dataset with 299 videos of 15-minute duration, annotated with bounding boxes and $80$ action classes at 1 FPS rate. The training and validation partitions contain 235 and 64 videos, amounting to 211k and 57k frames, respectively. We follow the standard evaluation protocol and report our results on the $60$-class subset of annotated actions~\cite{pan2021actor,zhao2021tuber,fan2021multiscale,feichtenhofer2019}. \textbf{UCF101-24}~\cite{soomro2012ucf101} contains 3207 untrimmed videos and contains box labels and annotations on a per-frame basis, with $24$ classes. We follow the standard protocol defined in ~\cite{singh2017online} and report our results on split-1. \textbf{JHMDB51-21}~\cite{jhuang2013towards} has 928 trimmed videos that are labelled with 21 action categories. We follow prior work and report the average results on the three splits. For all datasets, we report the standard mean average precision (mAP) computed considering as false positives all predictions corresponding to bounding boxes with IoU $<$ 0.5 w.r.t. a ground-truth box. \\

\noindent \textbf{Implementation details:} We initialize our model weights from the publicly available checkpoint of \cite{li2021improved} pretrained on Kinetics-400~\cite{kay2017kinetics}. The network architecture and the output sizes are summarized in the \cref{sec:arch}. The input to the model is $T=16$ frames sampled at a stride of $\tau=4$. All experiments are done using PyTorch~\cite{paszke2019pytorch}. We train our models using AdamW~\cite{loshchilov2017decoupled} with weight decay 0.0001, and set the trade-off hyperparameters in \cref{eq:total_loss} to $\lambda_{\alpha}=2.0$, $\lambda_{a}=6.0$ $\lambda_{L1}=5.0$, and $\lambda_{iou}=2.0$. We use a batch size of $16$ clips, and we train our models using 8 GeForce 3090 GPU cards. We train our model for 25 epochs with initial learning rate of 0.0001 and cosine decay. During training, we resize the videos to $256$ pixels without cropping. We add jitter to the ground-truth bounding boxes and apply color augmentation. We report our results using a single view with images directly resized to $256 \times 256$ pixels.

\section{Ablation studies}

\subsection{vs MViT + ROI align}
\label{ssec:mvit1}

\noindent We first show that our method offers significant improvement w.r.t. the standard two-stage approach of MViT. Because removing the last pooling stride of our model increases the number of FLOPs, we also train a model that preserves the original MViT structure. In such a scenario, the output is a volume of $t = 8$, $h = w = 8$ for an input resolution of $256 \times 256$. In the MViT + ROI align setting~\cite{li2021improved} a temporal pooling layer is added to remove the time dimension, and ROI align is used to extract the actor features from pre-computed bounding boxes. The actor-specific features are forwarded to a classifier to predict the class probabilities. In our setting, we generate a set of $64$ triplets consisting of bounding box coordinates with their corresponding actor likelihoods and class probabilities. We keep the predictions corresponding to the tokens for which the actor likelihood is over a threshold empirically set to be $\theta = 0.2$. We also consider reducing the FLOPs by working at a lower resolution and study how our method works at a resolution of $224$ pixels. The output number of tokens when working at $224$ resolution reduces to $14 \times 14$. We finally explore the scenario where the resolution is dropped to $224$, and the pooling layers are left as in the original MViTv2. Such a model produces only $7\times 7$ tokens at the central frame. The results are summarized in \cref{tab:mvit_roi}. 

\subsection{Token Selection}
\label{ssec:tokens}
\noindent As mentioned in \cref{ssec:ours}, the token selection design results in a different number of tokens to be assigned to the ground-truth. In this Section, we study how such selection affects performance. For an MViTv2-S without the last pooling layer, this results in $\tilde{L} = 2048$ tokens. In \cref{ssec:ours} we introduced two alternatives to the token selection designs, namely that of considering the central frame only for detection and temporal pooling for classification, which we refer to as \textbf{C+T}, and that depicted in \cref{fig:heads}, which we denote as \textbf{2(C+T)}. The former produces $\tilde{L} = 256$ tokens, whereas the latter produces $\tilde{L}=512$ tokens. In addition, we study three more alternatives, namely \textbf{singletons}, where we directly consider the $\tilde{L} = 2048$ tokens without further reduction; \textbf{tubelets}, whereby we directly apply temporal pooling to the output embeddings of the MViTv2-S backbone, producing the same $\tilde{L} = h \times w = 256$ for both the detection and recognition tasks; and \textbf{max-pooling of class predictions} which considers the central tokens for the actor detection task, and a max-pooling operation on the temporal domain over the outputs of the $\tilde{L} = 2048$ possible action tokens. The results in \cref{tab:token_Selection} indicate that a proper token selection is important to achieve a good tradeoff between detection and classification. We observe that two factors affect the most to improve the mAP: a high recall in the bounding box detection and a good selection of representatives for action tokens.

\subsection{Fixed vs Variable Aspect Ratio}
\label{ssec:additional_ablations}

\noindent Often, detection frameworks apply a scale augmentation by resizing the images to different scales, and by keeping the aspect ratio in the case of transformers. Such augmentation incurs a variable number of tokens $\tilde{L}$ to be assigned for each clip, which might affect the learning. We compare both approaches by training our model using a scale range between $240$ and $340$ pixels and keeping the aspect ratio. As shown in \cref{tab:ablation_alpha}, the performance drops significantly with respect to using a fixed number of tokens. We attribute this effect of matching a variable number of tokens to the ground-truth during training, which might require further hyperparameter optimization to improve convergence. We leave for future work improving the training in the case of variable-size images.

\subsection{Qualitative analysis}
\label{ssec:qual} 
\begin{figure}[t]
  \centering
 \includegraphics[width=0.48\linewidth]{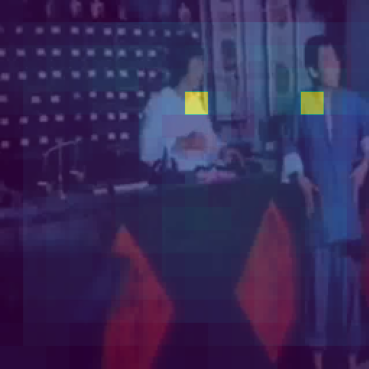}  \includegraphics[width=0.48\linewidth]{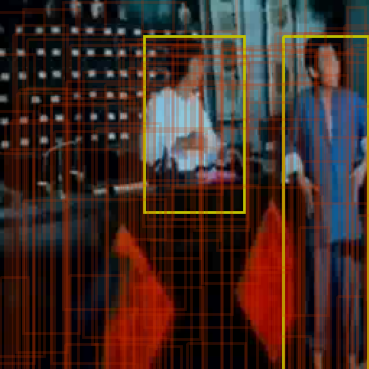} \\
 \includegraphics[width=0.48\linewidth]{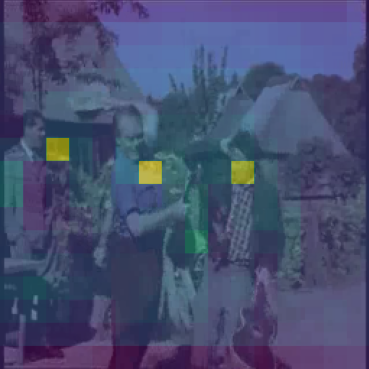}  \includegraphics[width=0.48\linewidth]{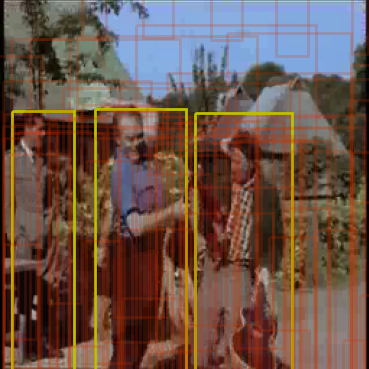}
 \\
 \includegraphics[width=0.48\linewidth]{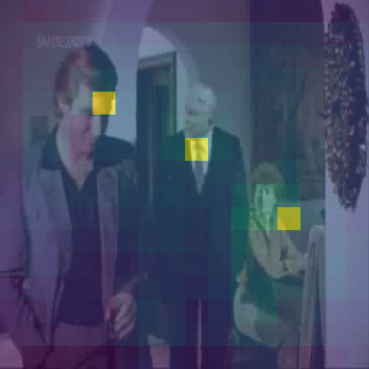}  \includegraphics[width=0.48\linewidth]{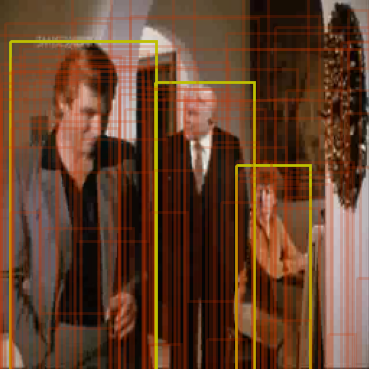}
\caption{\textbf{Qualitative analysis}. The images on the \textbf{left} show the confidence maps produced by the output $16\times 16$ spatial tokens (rescaled to the image size) w.r.t. the actor likelihood for the corresponding bounding box. For the sake of clarity, we only plot the $256$ tokens corresponding to one of the frames. The highlighted tokens are those selected as positive detections. The images on the \textbf{right} show all the bounding boxes computed by the corresponding tokens on the left. We overlay all the bounding boxes returned by each of the $16\times 16$ output tokens. In yellow we represent the bounding boxes corresponding to the confident tokens represented on the left. All other bounding boxes (in red) are assigned to the no-class label $\varnothing$, and are thus considered as negative predictions}
\label{fig:qual}
\vspace{-0.2in}
\end{figure}

\noindent In \cref{fig:qual} we show a visual demonstration of how the tokens carry the actor information properly. The left images show the confidence maps (i.e. $p(\alpha)$) for each of the $h \times w$ tokens at $t = \lfloor T/2 \rfloor$ (for the sake of clarity we illustrate only the confidence maps for the frame where the actor confidences were resulting in positive detections). We can see that the tokens around the actors in the frame are more confident than those that are farther away. In the right images, we overlap all the bounding boxes computed for the same $h \times w$ tokens, representing in yellow those corresponding to the activated tokens. We can see that while there are other bounding boxes around the two actors, only those that are maximally overlapping the ground-truth activate the actor likelihood with high confidence. We include more qualitative results in the \cref{sec:visual_analysis}.

\section{Comparison with State-of-the-art}
\subsection{Performance on AVA 2.2}
\input{ava.tex}

\label{ssec:ava22}
\noindent We compare our method against current state-of-the-art approaches in AVA 2.2, as summarized in \cref{tab:sota_ava}. We use symbols {\xmark} to denote methods relying on pre-computed bounding boxes, indicating a two-stage process, and symbols {\cmark} for those employing a single-stage approach. Additionally, we provide insights into the computational costs associated with each method. Note that two-stage methods listed in \cref{tab:sota_ava} utilize bounding boxes generated by \cite{feichtenhofer2019}, i.e. they operate a FasterRCNN-R101-FPN network that requires 246 GFLOPs on an input resolution of 512 pixels\cite{carion2020end}.

\noindent \textbf{Comparison with models pre-trained on K400:} We are mostly interested in attaining increased accuracy at a low computational cost. When comparing our method to two-stage approaches, we note that our method using MViTv2-$16\times4$ as the backbone surpasses all MViT methods, and achieves higher accuracy than MeMViT, with fewer FLOPs. 

\noindent \textbf{Comparison with recent single stage approaches:} We compare our method against TubeR~\cite{zhao2021tuber}, STMixer~\cite{wu2023stmixer} and EVAD~\cite{chen2023} which are the latest works on single-stage action detector. For a fair comparison, and given that we have trained on models initialized on K400, we do not report and compare large models pre-trained on K700. In K400 we achieve $30.0$ mAP comparable to TubeR and STMixer. However, our method offers a simpler solution without explicit context modeling and without the use of a decoder. 

\noindent \textbf{Generalization} To further demonstrate the capabilities of our method, we trained the same ViT-B backbone used for EVAD~\cite{chen2023}, pre-trained on K400 and VideoMAE~\cite{tong2022videomae}, without token dropout. Following EVAD, we trained our model using a resolution of $288$ pixels. For this model, we used only the central tokens for both actor detection and action classification, resulting in $\tilde{L} = 18\times18 = 324$ output tokens. We added a cross attention layer to the action classification head to account for the temporal information, whereby the $\tilde{L}$ output tokens first attend all the $8 \times 18 \times 18 = 2592$ spatio-temporal tokens before being forwarded to the action classification MLP. Note that this cross-attention layer (with $324$ queries and $2592$ keys and values), amounts to only $4.7$ GFLOPs, maintaining a computationally efficient action head. The results in \cref{tab:sota_ava} (bottom) show that our method offers a similar complexity/performance tradeoff w.r.t. EVAD.

\noindent \textbf{Scaling up to large backbones:}
While the scope of our proposed approach is to develop efficient methods for training a video transformer for action localization, we explore how our proposed approach scales to larger models. To this end, we train a Hiera-L~\cite{ryali2023hiera} in an end-to-end fashion using the publicly available checkpoint pre-trained on Kinetics400, which was first pre-trained in a self-supervised manner using Masked AutoEncoders~\cite{MaskedAutoencoders2021}. We directly reuse our training recipe to train the large model, without any further parameter optimization. We observe that our approach competes with the two-stage approach of Hiera-L~\cite{ryali2023hiera}, although results in sub-par performance. While Hiera-L obtains $39.8$ mAP in a two-stage approach incurring in $413+246$ GFLOPs, our single-stage equivalent obtains $38.5$ mAP with $650$ GFLOPs (full results are included in the \cref{sec:larger}). We attribute this sub-par performance to the difficulty of training large models in an end-to-end fashion. We leave for future research the optimization of big models in our setting.  

\subsection{Additional Datasets}
\noindent To demonstrate the effectiveness of our approach we evaluate our method on \textbf{UCF101-24}~\cite{soomro2012ucf101} and \textbf{JHMDB51-21}~\cite{jhuang2013towards} datasets. Results of our approach are shown in~\cref{table:ucf_sota}. We note that compared to EVAD~\cite{chen2023} we have similar performance with a backbone of much smaller capacity, i.e. our backbone has $121$ GFLOPs while EVAD~\cite{chen2023} has $243$ GFLOPS. Our approach surpasses EVAD on UCF24, and lies behind in JHMDB. However, for JHMDB, we observed that our model attains precision and recall of of $95\%$ and $97\%$ correspondingly, indicating that further attention needs to be put towards improving accuracy.

\setlength{\tabcolsep}{1pt}
\begin{table}[t!]
\small
\begin{center}
    \begin{minipage}[t]{1\linewidth}
        \centering
        
        \begin{subtable}[t]{1.\linewidth}
            \begin{tabular}{lcccccc} 
                \shline Method & End-to-end & Backbone & JHMDB & UCF24  \\
                \hline
                ACAR-Net~\cite{pan2021actor} & \xmark & SF-R50 & - & $84.3$ \\
                ACT~\cite{kalogeiton2017action} & \cmark & VGG & $65.7$ & $69.5$  \\
                MOC~\cite{li2020ava}$^\ast$ & \cmark  & DLA32 & $70.8$ & $78.0$ \\
                ACRN~\cite{sun2018actor} & \cmark & S3D-G & $77.9$ & -  \\
                YOWO~\cite{kopuklu2019yowo} & \cmark  & 3D-X101 & $80.4$ & $74.4$  \\
                WOO~\cite{chen2021watch} & \cmark  & SF-R101-NL & $80.5$ & -  \\
                TubeR~\cite{zhao2021tuber}$^\ast$ & \cmark & I3D & $80.7$ & $81.3$  \\
                TubeR~\cite{zhao2021tuber} & \cmark & CSN-152 & - & $83.2$ \\
                STMixer~\cite{wu2023stmixer} & \cmark  & SF-R101-NL & $\bf{86.7}$ & $83.7$ \\
                \rowcolor{defaultcolor}
                 Ours (BMViT) & \cmark & MViTv2-S & 80.7 & $\bf{85.6}$ \\ \midrule
                 
                                 EVAD~\cite{chen2023} @ 288 & \cmark  & ViT-B & $\bf{90.2}$ & $85.1$ \\
\rowcolor{defaultcolor}
                 Ours (BMViT) @ 288 & \cmark & ViT-B & 85.4 & $\bf{87.3}$ \\
                 \rowcolor{defaultcolor}
                 Ours (BMViT) @ 288 & \cmark  & ViT-B$^\dag$ & 88.4 & $\bf{90.7}$ \\
                \shline
            \end{tabular}
            
        \end{subtable}
    \end{minipage}
    \vspace{-2mm}
    \caption{\label{table:ucf_sota} \textbf{Comparison with the state-of-the-art on UCF101-24 and JHMDB.} \Checkmark denotes an end-to-end approach using a unified backbone, and \XSolidBrush denotes the use of two separated backbones, one of which is Faster R-CNN-R101-FPN (246 GFLOPs~\cite{ren2015faster}) to pre-compute person proposals.  $T \times \tau$ refers to the frame number and corresponding sample rate. Methods marked with $^\ast$ leverage optical flow input.$^\dag$: ViT-B backbone pre-trained on K710 and fine-tuned on K400 from VideoMAE V2~\cite{wang2023videomae}.}
\end{center}
\vspace{-7mm}
\end{table}

%% file: ablation.tex
\begin{table*}[t!]

    \begin{subtable}[b]{0.38\linewidth}
        \centering
        \footnotesize
        \renewcommand{\arraystretch}{1.2}
        \setlength{\tabcolsep}{1.5pt} %
        \begin{tabular*}{\linewidth}{@{\extracolsep{\fill}\;}lcc@{\extracolsep{\fill}\;}}
            \toprule
            \multicolumn{1}{c}{{\bf Setting} ($t \times hw$)} & {\bf GFLOPs}&  {\bf mAP}  \\
            \midrule
                MViTv2-S~\cite{li2021improved} & 64 + 246 & 26.8 \\
                Inp: $16\times 224^2$, Out: $8\times 7^2 $ & \bf{65} & 26.8 \\
                Inp: $16\times 224^2$, Out: $8\times 14^2 $ & 87.9 & 27.4 \\
                Inp: $16\times 256^2$, Out: $8\times 8^2 $ & 90.7 & 27.5 \\
                Inp: $16\times 256^2$, Out: $8\times 16^2 $ & 121.2 & \bf{30.0} \\
            
            \bottomrule
        \end{tabular*}
        \caption{vs MViT + ROI align}
        \label{tab:mvit_roi}
        \vspace{0.05in}
    \end{subtable}
    \hfill
    \begin{subtable}[b]{0.30\linewidth}
        \centering
        \footnotesize
        \renewcommand{\arraystretch}{1.2}
        \setlength{\tabcolsep}{0pt} 
        \begin{tabular*}{\linewidth}{@{\extracolsep{\fill}\;}cccc@{\extracolsep{\fill}\;}}
            \toprule
            \textbf{Method} & \textbf{mAP} & \textbf{Prec.} & \textbf{Recall}\\
            \midrule
            Singleton & 28.5 & 84.2 & 90.4 \\
            Tubelet & 28.7 & 83.2 & 90.5\\
            C + T & 29.1 & 81.1 & 92.7\\
            Max Pooling & 28.3 & 84.1 & 91.3 \\
        2(C+T) & \textbf{30.0} & 80.0 & 92.0\\
            \bottomrule
        \end{tabular*}
        \caption{Token Selection}
        \label{tab:token_Selection}
        \vspace{0.05in}
    \end{subtable}
    \hfill
    \begin{subtable}[b]{0.22\linewidth}
        \centering
         \footnotesize
         \renewcommand{\arraystretch}{1.2}
         \setlength{\tabcolsep}{0pt} %
         \begin{tabular*}{\linewidth}{@{\extracolsep{\fill}\;}lc@{\extracolsep{\fill}\;}}
            \toprule
            \multicolumn{1}{c}{\textbf{Method}} & \textbf{mAP} \\
             \midrule
                    Variable aspect ratio & 28.5 \\
                    Fixed aspect ratio & \textbf{30.0} \\
             \bottomrule
         \end{tabular*}
         \caption{Impact of aspect ratio}
         \label{tab:ablation_alpha}
         \vspace{0.05in}
    \end{subtable}

    \hfill   \vspace{-7pt} 
    \caption{ \textbf{Ablation studies on AVA 2.2} All experiments are done using an MViTv2-S~\cite{li2021improved} pre-trained on K400. \textbf{a)} We study the relation between complexity and input/output resolutions by removing or not the last pooling stride. Note that MViTv2-S~\cite{li2021improved} requires external bounding boxes, reportedly adding 246 GFLOPs to the overall inference. \textbf{b)} We study the impact of different methods for token selection to perform the bipartite matching. \textbf{c)} We study whether keeping the input aspect ratio affects the training, observing that a variable number of tokens results in difficult convergence.}
    \vspace{-0.1in}
\end{table*}

%% file: ava.tex
\begin{table*}[t]
	\vspace{-0.5\baselineskip}
	\centering
	
	\renewcommand{\arraystretch}{0.92}
	\scalebox{1.0}{
            \begin{tabular}{c c c c c c c c c}
			\toprule
			\textbf{Method} & \textbf{Pretraining}  & \textbf{mAP} & \textbf{GFLOPs} & \textbf{Res.} & \textbf{Backbone} & \textbf{End-to-end} \\ 
			\midrule
			ACAR-Net~\cite{pan2021actor}        & K400          & $28.8$    & $205+246$  & 256 & SF-R50-NL & \xmark  \\ 
   			MViTv1-B~\cite{fan2021multiscale}& K400       & $27.3$ & $455 + 246$  & 224 &  MViTv1-B     & \xmark \\ 
   			MViTv2-S~\cite{li2021improved}   & K400       & $26.8$ & $65  + 246$ & 224 &    MViTv2-S   & \xmark \\ 
			MViTv2-B~\cite{li2021improved}   & K400       & $28.1$ & $225 + 246$ & 224 &   MViTv2-B   & \xmark \\ 
                MeMViT~\cite{wu2022}             & K400       & $28.5$ & $59  + 246$ & 224 &      MViTv1  & \xmark \\
                WOO~\cite{chen2021watch}         & K400       & $25.4$    & $148$ & 256 &  SF-R50  & \cmark \vspace{0.3ex} \\
                STMixer~\cite{wu2023stmixer}     & K400     & $27.8$    & N/A & 256 & SF-R50 & \cmark 
                \vspace{0.3ex} \\
                \rowcolor{defaultcolor}
                Ours (BMViT)                           & K400 & $\bf{30.0}$ & \textbf{121} & 256 &          MViTv2-S $16\times 4$          & \cmark \vspace{0.3ex} \\
\midrule
   			VideoMAE~\cite{tong2022videomae}   & K400, MAE      & $31.8$ & $180  + 246$ & 224 &    ViT-B  & \xmark \\ 
            EVAD~\cite{chen2023}  & K400, MAE&  $\bf{32.1}$ & $425$ & 288 &  ViT-B& \cmark \vspace{0.3ex} \\   
                
                    \rowcolor{defaultcolor}
                Ours (BMViT)                           & K400, MAE & $31.4$ & \textbf{350} & 288 &          ViT-B $16\times 4$          & \cmark \vspace{0.3ex} \\
 
   \bottomrule
		\end{tabular}} 
  \caption{Comparison w.r.t. state-of-the-art (reported with mean Average Precision; mAP~$\uparrow$) on AVA v2.2~\cite{gu2018}. “Res.” denotes frame resolution. }
	\vspace{-0.7\baselineskip}
	\label{tab:sota_ava}
\end{table*}

%% file: conclusion.tex
\section{Conclusion}
\label{sec:conclusion}
\noindent In this paper, we presented a simple method for the direct training of Vision Transformers for end-to-end action localization. We showed that the output  tokens of a vision transformer can be independently forwarded to corresponding MLP heads to have a fixed sequence of predictions similar to DETR. Using a bipartite matching loss it is possible to train the backbone directly to perform both tasks without compromising performance. Our results show that simple models achieve similar accuracy to equivalent two-stage approaches.%

\paragraph{Acknowledgments} Ioanna Ntinou is funded by Queen Mary Principal's PhD Studentships. This research utilized Queen Mary's Apocrita HPC facility, supported by QMUL Research-IT. \href{http://doi.org/10.5281/zenodo.438045}{http://doi.org/10.5281/zenodo.438045}

%% file: appendix.tex
\appendix
\onecolumn
\begin{appendices}

\section{Architecture details}
\label{sec:arch}
\noindent Herein, we provide the architecture details of our approach when using both the MViTv2-S~\cite{li2021improved} and ViT-B~\cite{tong2022videomae} as backbones. Please refer to \cite{li2021improved}, \cite{tong2022videomae} and to \href{https://github.com/IoannaNti/BMViT}{https://github.com/IoannaNti/BMViT} for full implementation details.
\input{arch2}

\begin{figure*}[b!]
  \centering
 \includegraphics[width=0.97\linewidth]{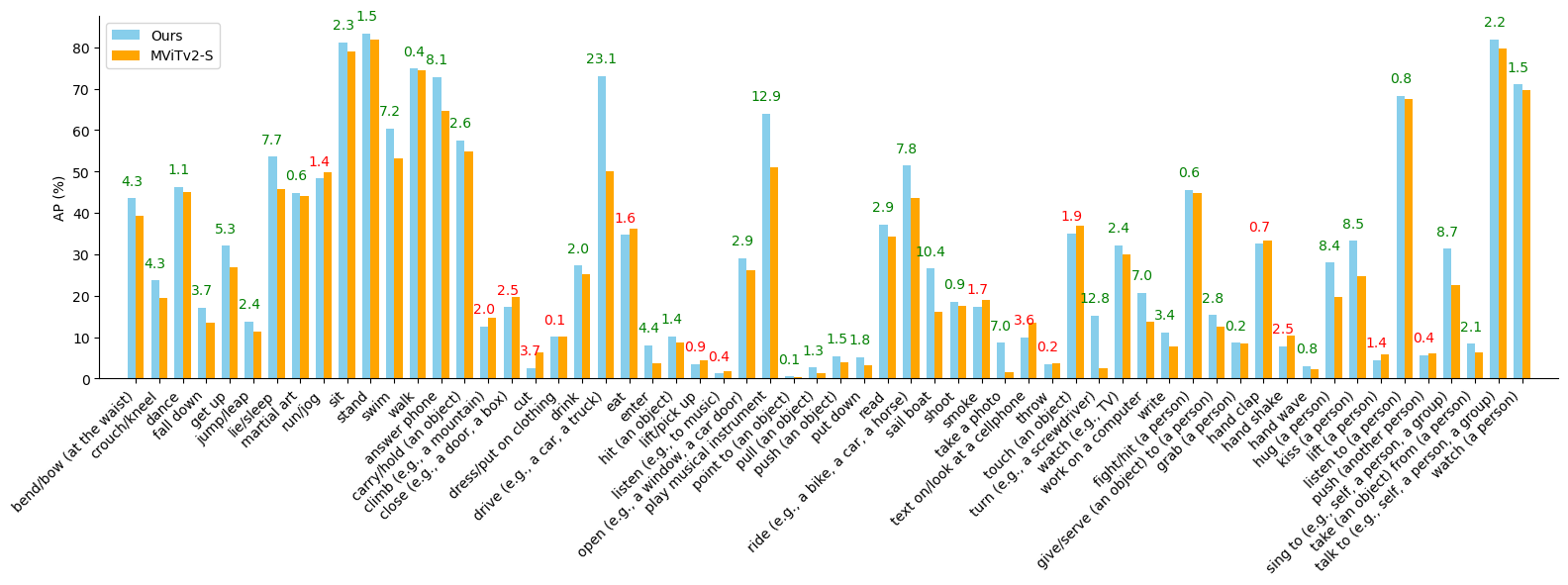}
\caption{Per-category AP for \textbf{Our} single stage action detection method (30.0 mAP) and \textbf{MViTv2-S} (27.0 mAP) on AVA v.2. On top of the bar there is the difference per-class where 
categories with increased accuracy are marked in \textbf{\textcolor[RGB]{0,255,0}{green}} and those decreased with our method in  \textbf{\textcolor[RGB]{255,0,0}{red}}.}
\label{fig:bar}
\vspace{-0.2in}
\end{figure*}

\section{Per-class analysis}
\label{sec:PCA}
\noindent In \cref{fig:bar} we present the performance per-category of our single-stage model built on MViTv2-S~\cite{li2021improved} and the corresponding two-stage approach of the same MViT on AVAv2.2~\cite{li2020ava}. To compute the per-class accuracy, we re-trained the backbone using the same settings as in \cite{li2021improved}, obtaining $27.05$ mAP, which is indeed $+0.2$ w.r.t. the reported results. Our approach demonstrates improvements in $44$ out of the $60$ categories, notably increasing in categories like `drive (e.g., a car, a truck)' with an impressive $+23.1$ mAP increase and `turn (e.g., a screwdriver)' with a significant $+12.8$ mAP boost.  Intriguingly, the performance trends across categories remain consistent between our method and the two-stage MViT, suggesting the feasibility of employing the same representation for both actor localization and action detection tasks.

\section{Additional results}
\label{sec:larger}

\noindent \textbf{Larger backbones} As mentioned in the main document, we trained our method using a Hiera-L backbone. The results of our method against large-scale state-of-the-art methods are shown in \cref{tab:sota_ava_large}. We observe that while our method falls short w.r.t. the two-stage counterpart, our simple, single-stage approach delivers competitive results. 

\noindent \textbf{Additional bounding boxes} In \cref{eq:total_loss} in the main document, the training objective considers the bounding box error w.r.t. only the annotated bounding boxes. We want to note that because we distinguish between an actor and a person, the bounding box loss could be also backpropagated w.r.t. bounding boxes corresponding to no-actors, if available. This could improve the precision of the detections and as such the overall performance. We conducted an experiment to study the impact of such approach, observing no effect (neither positive nor negative), in the results. We attribute this to the fact that the number of bounding boxes that correspond to a no-actor on AVA 2.2. is rather small compared to those of the annotated actors. 

\section{Visual analysis}
\label{sec:visual_analysis}
\begin{figure*}[t!]
  \centering
 \includegraphics[width=1\linewidth]{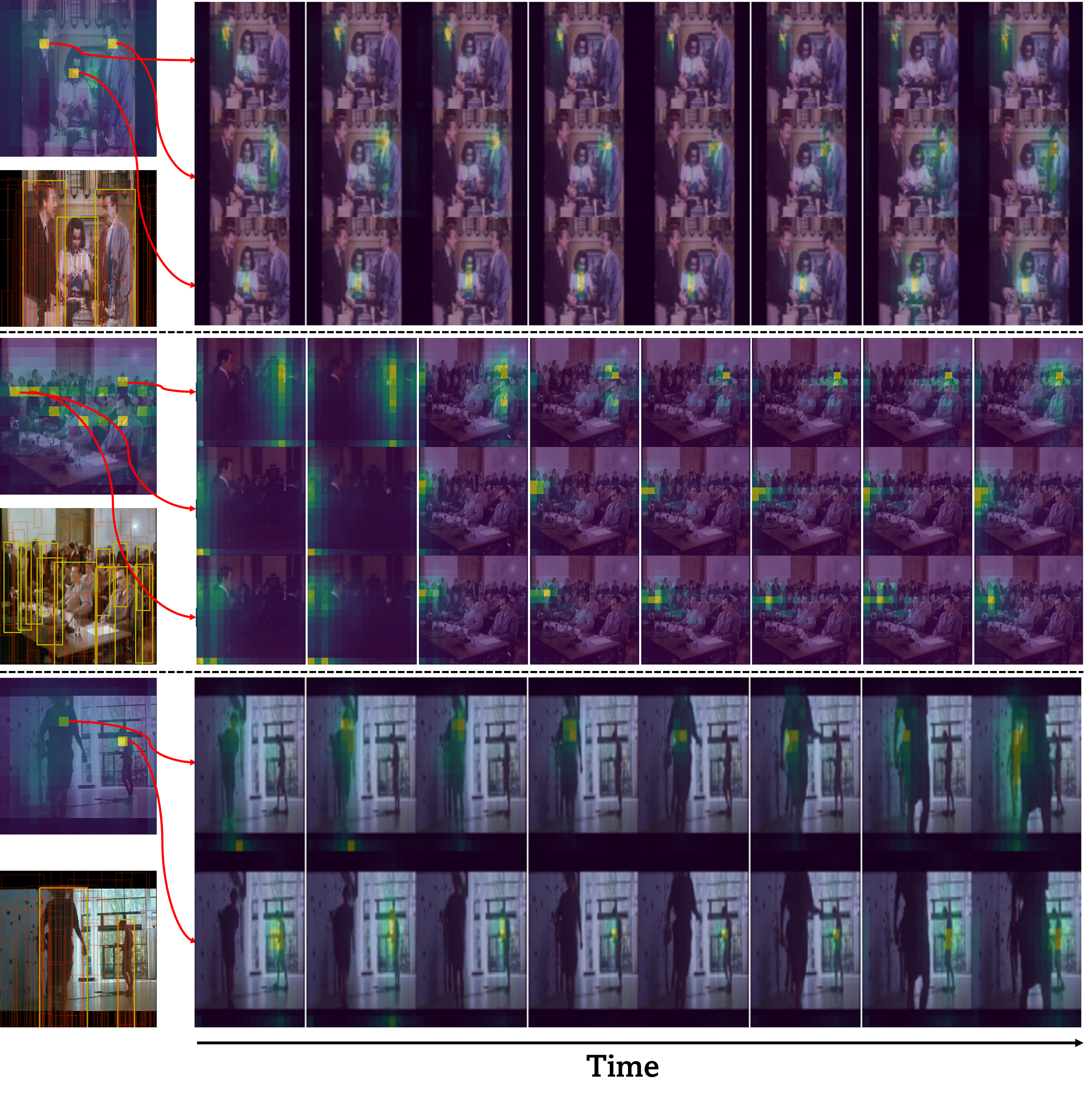}
\caption{\textbf{Qualitative analysis} (better seen in color and zoomed in). We provide three qualitative examples from three corresponding validation videos from AVA2.2. On the \textbf{left top} image we represent the confidence scores $p(\alpha)$ for the actor-no actor prediction, for each of the $16\times16$ output tokens corresponding to one of the central frames. Those with high confidence $p(\alpha) > \theta$ are selected as positive examples, and their corresponding bounding boxes and class predictions will then form the final outputs. The images in the \textbf{left bottom} are the bounding boxes predicted by each of the same $16\times 16$ output tokens, with those in yellow corresponding to the positive tokens (i.e. to the final output bounding boxes). On the \textbf{right} we represent the last layer's attention maps corresponding to each of the selected tokens in the left, i.e. their attention scores w.r.t. the whole $8 \times 16 \times 16$ spatio-temporal tokens. We observe that the confident tokens not only attend to the central information to produce the bounding box, but also track the corresponding actor across the video to estimate the corresponding actions. In the second example, we only represent three actors for the sake of clarity. We observe that even with a change of scene, the attention maps can properly track each actor's information. In the last example, the self-attention maps show how they can track each actor's despite the self-occlusion. These examples show that our method can track the actor's information and regress the bounding boxes with a Vision Transformer that assigns each vision token a different output, which are assigned to the ground-truth set through bipartite matching. }
\label{fig:qualAppendix}
\vspace{-0.2in}
\end{figure*}

\noindent In \cref{fig:qualAppendix} we show three visual examples complementing those presented in \cref{fig:qual} in the main paper. In these examples, we also show the attention maps for each of the tokens that are chosen to represent an actor with $p(\alpha) > \theta$. We can see that the attention maps show how each token is indeed tracking the actor for which they carry the corresponding information, illustrating how our approach can enable the visual tokens to carry bounding box information regarding the central frame, as well as the class information that requires spatio-temporal reasoning.

\input{ava_large}

\twocolumn

\end{appendices}

%% file: arch2.tex
\newcommand{\blocks}[3]{\multirow{3}{*}{\(\left[\begin{array}{c}\text{1$\times$1$^\text{2}$, #2}\\[-.1em] \text{1$\times$3$^\text{2}$, #2}\\[-.1em] \text{1$\times$1$^\text{2}$, #1}\end{array}\right]\)$\times$#3}
}
\newcommand{\blocket}[4]{\multirow{3}{*}{\(\left[\begin{array}{c}\text{1$\times$1$^\text{2}$, #1}\\[-.1em] \text{$3$$\times$3$^\text{2}$, #2}\\[-.1em] \text{1$\times$1$^\text{2}$, #3}\end{array}\right]\)$\times$#4}
}

\newcommand{\blockatta}[3]{\multirow{2}{*}{\(\left[\begin{array}{c}\text{\eicolor{MHPA}(\wcolor{#1})}\\[-.1em] \text{MLP(\wcolor{#2})}\end{array}\right]\)$\times$#3}
}

\newcommand{\blockatts}[3]{\multirow{2}{*}{\(\left[\begin{array}{c}\text{\eicolor{MHSA}(\wcolor{#1})}\\[-.1em] \text{MLP(\wcolor{#2})}\end{array}\right]\)$\times$#3}
}

\newcommand{\blockattc}[3]{\multirow{2}{*}{\(\left[\begin{array}{c}\text{\eicolor{MHCA}(\wcolor{#1})}\\[-.1em] \text{MLP(\wcolor{#2})}\end{array}\right]\)$\times$#3}
}

\newcommand{\blockt}[3]{\multirow{3}{*}{\(\left[\begin{array}{c}\text{\underline{3$\times$1$^\text{2}$}, #2}\\[-.1em] \text{1$\times$3$^\text{2}$, #2}\\[-.1em] \text{1$\times$1$^\text{2}$, #1}\end{array}\right]\)$\times$#3}
}
\newcommand{\outsizes}[7]{\multirow{#7}{*}{\(\begin{array}{c} \text{\emph{Slow}}: \text{#1$\times$#2$^\text{2}$}\\[-.1em] \text{\emph{Fast}}: \text{#4$\times$#5$^\text{2}$}\end{array}\)}
}

\newcommand{\outsizesRaw}[4]{\multirow{#4}{*}{\(\begin{array}{c}  \text{#1$\times$#2\x #3}\\[-.1em]  \end{array}\)}}
\newcommand{\outsizesRawD}[5]{\multirow{#5}{*}{\(\begin{array}{c}  \text{#1\x#2\x#3\x#4}\\[-.1em]  \end{array}\)}}

\newcommand{\outsizesGamma}[4]{\multirow{#4}{*}{\(\begin{array}{c}  \text{#1\gat\x (#2\gaxy)$^2$}\\[-.1em]  \end{array}\)}}

\newcommand{\outsizesSF}[5]{\multirow{#5}{*}{\(\begin{array}{cc} \text{\emph{Slow}}:& \text{#1$\times$ #2$^\text{2}$}\\[-.1em] \text{\emph{Fast}}:& \text{#3$\times$#4$^\text{2}$}\end{array}\)}}

\begin{figure}[htbp]
    \centering
    \begin{subfigure}[t]{0.5\textwidth}
        \centering
		\begin{tabular}{c|c|c}
			stage & operators & output sizes \\
			\shline
			\multirow{1}{*}{data} & \multirow{1}{*}{stride \tcolor{4}\x1\x1}   &  \outsizesRaw{\tcolor{16}}{\xycolor{256}}{\xycolor{256}}{1}   \\
			\hline
			
			\multirow{2}{*}{cube$_1$} & \multicolumn{1}{c|}{3\x7\x7, {96}} &    \outsizesRawD{\wcolor{$96$}}{\tcolor{8}}{\xycolor{64}}{\xycolor{64}}{2}    \\
			& stride 2\x4\x4   \\
			\hline
			\multirow{2}{*}{scale$_2$}  & \blockatta{96}{{384}}{1} & \outsizesRawD{\wcolor{$96$}}{\tcolor{8}}{\xycolor{64}}{\xycolor{64}}{2}  \\
			&  & \\
			\hline
			\multirow{2}{*}{scale$_3$}  & \blockatta{192}{{768}}{2} & \outsizesRawD{\wcolor{$192$}}{\tcolor{8}}{\xycolor{32}}{\xycolor{32}}{2}  \\
			&  & \\
			\hline
			\multirow{2}{*}{scale$_4$}  & \blockatta{384}{{1536}}{11} & \outsizesRawD{\wcolor{$384$}}{\tcolor{8}}{\xycolor{16}}{\xycolor{16}}{2}  \\
			&  & \\
			\hline
			\multirow{2}{*}{scale$_5$}  & \blockatta{768}{{3072}}{2} & \outsizesRawD{\wcolor{$768$}}{\tcolor{8}}{\xycolor{16}}{\xycolor{16}}{2}  \\
			&  & \\
			\hline
            \multirow{1}{*} {proj} & \multicolumn{1}{c|}{1\x1\x1, {512}} & \outsizesRaw{\wcolor{$512$}}{$8$}{$256$}{1} \\
            \hline
            \multirow{3}{*} {head} &  \multirow{3}{*} { $\left\{\begin{matrix} \text{MLP}(\wcolor{512}) \\ \text{MLP}(\wcolor{512})  \\ \text{MLP}(\wcolor{512})\end{matrix} \right.$ } & \multirow{3}{*} {$\left\{\begin{matrix} \xycolor{512} \times \wcolor{4} \\ \xycolor{512} \times \wcolor{2} \\ \xycolor{512} \times \wcolor{C} \end{matrix} \right.$ } \\ & & \\ & & \\
            \hline   
		\end{tabular}
	\vspace{.1em}
	\caption{ Our pipeline with MViTv2-S as bockobone.
	}
	\label{tab:arch_instances_mvit}
    \end{subfigure}%
    \begin{subfigure}[t]{0.5\textwidth}
        \centering
		\begin{tabular}{c|c|c}
			stage & operators & output sizes \\
			\shline
			\multirow{1}{*}{data} & \multirow{1}{*}{stride \tcolor{4}\x1\x1}   &  \outsizesRaw{\tcolor{16}}{\xycolor{288}}{\xycolor{288}}{1}   \\
			\hline
			
			 \multirow{2}{*}{cube$_1$} & \multicolumn{1}{c|}{3\x16\x16, {768}} &    \outsizesRawD{\wcolor{$768$}}{\tcolor{8}}{\xycolor{18}}{\xycolor{18}}{2}    \\
			 & stride 2\x16\x16   \\
			 \hline
			\multirow{2}{*}{cube$_2$}  & \blockatts{768}{{3072}}{12} & \outsizesRawD{\wcolor{$768$}}{\tcolor{8}}{\xycolor{18}}{\xycolor{18}}{2}  \\
		&  & \\
          \hline

           \multirow{5}{*} {head} &  \multirow{5}{*} { $\left\{\begin{matrix} \text{MLP}(\wcolor{768}) \\ \text{MLP}(\wcolor{768})  \\ \left\{\begin{matrix} \text{\eicolor{MHCA}}(\wcolor{768}) \\ \text{MLP} (\wcolor{3072}) \\ \text{MLP} (\wcolor{768}) \end{matrix} \right.   \end{matrix} \right.$ } & \multirow{5}{*} {$\left\{\begin{matrix} \xycolor{768} \times \wcolor{4} \\ \xycolor{768} \times \wcolor{2} \\  \xycolor{768} \times \wcolor{C} \end{matrix}  \right.$ } \\ & & \\ & & \\ & & \\ & & \\
            \hline   
		\end{tabular}
	\vspace{.1em}
	\caption{Our pipeline with ViT-B-S as backobone.	}
	\label{tab:arch_instances_vitb}
    \end{subfigure}
    \caption{\ref{tab:arch_instances_mvit} The network architecture resembles that of MViTv2-S~\cite{li2021improved}, with the pooling layer after scale$_4$ removed. The output features are projected to $512$ dimensions and forwarded to three parallel heads that predict for each token the bounding box coordinates, the probability of the bounding box being an actor, and the class predictions. \ref{tab:arch_instances_vitb}
    The network architecture resembles that of ViT-B~\cite{li2021improved}. The output tokens corresponding to $t=\lfloor T/2 \rfloor$ are forwarded to two parallel heads that predict for each token the bounding box coordinates and the probability of the bounding box being an actor. For the class prediction, we apply cross-attention between all output tokens of shape $8 \times 18 \times 18$  and the ones corresponding to the central frame. The attended tokens are then passed through an MLP for class predictions. }
    \label{fig:both_tables} \vspace{-3mm}
\end{figure}

%% file: ava_large.tex
\begin{table*}[t]
	\vspace{-0.5\baselineskip}
	\centering
	
	\renewcommand{\arraystretch}{0.92}
	\scalebox{0.9}{
            \begin{tabular}{c c c c c c c c c}
			\toprule
			\textbf{Method} & \textbf{Pretraining}  & \textbf{mAP} & \textbf{GFLOPs} & \textbf{Res.} & \textbf{Backbone} & \textbf{End-to-end} \\ 
			\midrule
            VideoMAE~\cite{tong2022videomae}    & K400, VideoMAE        & $31.8$ & 180+246 & 224 & ViT-B & \xmark \\
            VideoMAE~\cite{tong2022videomae}    & K400, VideoMAE        & $37.0$ & 597+246 & 224 & ViT-L & \xmark \\
            UMT~\cite{li2023unmasked}        & K400, UMT & $32.7$ & 180+246 & 224 &    ViT-B                     & \xmark \vspace{0.3ex} \\
	    UMT~\cite{li2023unmasked}        & K400, UMT &$39.0$ & 596+246 & 224 &    ViT-L                     & \xmark \vspace{0.3ex} \\
            Hiera-L~\cite{ryali2023hiera}      & K400, MAE  & $\bf{39.8}$ & 413+246 & 224  &  Hiera-L & \xmark \vspace{0.3ex} \\
           TubeR~\cite{zhao2021tuber}       & K400, IG65M~\cite{ig65} &$29.2$ & 97& 256&   CSN-50    & \cmark \vspace{0.3ex} \\
           TubeR~\cite{zhao2021tuber}          & K400, IG65M~\cite{ig65}   & $33.4$   & $138$ & 256& CSN-152& \cmark \vspace{0.3ex} \\
            EVAD~\cite{chen2023}             & K400, VideoMAE&  $32.3$ & $243$ & 288 &  ViT-B& \cmark \vspace{0.3ex} \\
            STMixer~\cite{wu2023stmixer}        & K400, VideoMAE   & $32.6$    & N/A & 256 & ViT-B & \cmark \vspace{0.3ex} \\
            \rowcolor{defaultcolor}
            Ours                        & K400, MAE &   $38.5$ & 650 & 256 & Hiera-L & \cmark \vspace{0.3ex} \\            
   \bottomrule
		\end{tabular}} 
  \caption{Comparison w.r.t. state-of-the-art (mean Average Precision; mAP~$\uparrow$) on AVA v2.2~\cite{gu2018}. “Res.” denotes frame resolution. }
	\vspace{-0.5\baselineskip}
	\label{tab:sota_ava_large}
\end{table*}

%% file: PaperForReview.bbl
\begin{thebibliography}{10}\itemsep=-1pt

\bibitem{carion2020end}
Nicolas Carion, Francisco Massa, Gabriel Synnaeve, Nicolas Usunier, Alexander Kirillov, and Sergey Zagoruyko.
\newblock End-to-end object detection with transformers.
\newblock In {\em ECCV}, 2020.

\bibitem{chao2018}
Yu{-}Wei Chao, Sudheendra Vijayanarasimhan, Bryan Seybold, David~A. Ross, Jia Deng, and Rahul Sukthankar.
\newblock Rethinking the faster {R-CNN} architecture for temporal action localization.
\newblock In {\em CVPR}, 2018.

\bibitem{chen2023}
Lei Chen, Zhan Tong, Yibing Song, Gangshan Wu, and Limin Wang.
\newblock Efficient video action detection with token dropout and context refinement.
\newblock In {\em ICCV}, 2023.

\bibitem{chen2021watch}
Shoufa Chen, Peize Sun, Enze Xie, Chongjian Ge, Jiannan Wu, Lan Ma, Jiajun Shen, and Ping Luo.
\newblock Watch only once: An end-to-end video action detection framework.
\newblock In {\em ICCV}, 2021.

\bibitem{dosovitskiy2020vit}
Alexey Dosovitskiy, Lucas Beyer, Alexander Kolesnikov, Dirk Weissenborn, Xiaohua Zhai, Thomas Unterthiner, Mostafa Dehghani, Matthias Minderer, Georg Heigold, Sylvain Gelly, Jakob Uszkoreit, and Neil Houlsby.
\newblock An image is worth 16x16 words: Transformers for image recognition at scale.
\newblock In {\em ICLR}, 2021.

\bibitem{fan2021multiscale}
Haoqi Fan, Bo Xiong, Karttikeya Mangalam, Yanghao Li, Zhicheng Yan, Jitendra Malik, and Christoph Feichtenhofer.
\newblock Multiscale vision transformers.
\newblock In {\em ICCV}, 2021.

\bibitem{faure2022}
Gueter~Josmy Faure, Min-Hung Chen, and Shang-Hong Lai.
\newblock Holistic interaction transformer network for action detection.
\newblock In {\em WACV}, 2023.

\bibitem{feichtenhofer2019}
Christoph Feichtenhofer, Haoqi Fan, Jitendra Malik, and Kaiming He.
\newblock Slowfast networks for video recognition.
\newblock In {\em ICCV}, 2019.

\bibitem{ig65}
Deepti Ghadiyaram, Matt Feiszli, Du Tran, Heng~Wang Xueting~Yan, and Dhruv Mahajan.
\newblock Large-scale weaklysupervised pre-training for video action recognition.
\newblock In {\em CVPR}, 2019.

\bibitem{girdhar2019video}
Rohit Girdhar, Joao Carreira, Carl Doersch, and Andrew Zisserman.
\newblock Video action transformer network.
\newblock In {\em CVPR}, 2019.

\bibitem{gu2018}
Chunhui Gu, Chen Sun, David~A. Ross, Carl Vondrick, Caroline Pantofaru, Yeqing Li, Sudheendra Vijayanarasimhan, George Toderici, Susanna Ricco, Rahul Sukthankar, Cordelia Schmid, and Jitendra Malik.
\newblock {AVA:} {A} video dataset of spatio-temporally localized atomic visual actions.
\newblock In {\em CVPR}, 2018.

\bibitem{MaskedAutoencoders2021}
Kaiming He, Xinlei Chen, Saining Xie, Yanghao Li, Piotr Doll{\'a}r, and Ross Girshick.
\newblock Masked autoencoders are scalable vision learners.
\newblock In {\em CVPR}, 2021.

\bibitem{he2017}
Kaiming He, Georgia Gkioxari, Piotr Dollar, and Ross Girshick.
\newblock Mask r-cnn.
\newblock In {\em ICCV}, 2017.

\bibitem{jhuang2013towards}
Hueihan Jhuang, Juergen Gall, Silvia Zuffi, Cordelia Schmid, and Michael~J Black.
\newblock Towards understanding action recognition.
\newblock In {\em ICCV}, 2013.

\bibitem{kalogeiton2017action}
Vicky Kalogeiton, Philippe Weinzaepfel, Vittorio Ferrari, and Cordelia Schmid.
\newblock Action tubelet detector for spatio-temporal action localization.
\newblock In {\em ICCV}, 2017.

\bibitem{kay2017kinetics}
Will Kay, Joao Carreira, Karen Simonyan, Brian Zhang, Chloe Hillier, Sudheendra Vijayanarasimhan, Fabio Viola, Tim Green, Trevor Back, Paul Natsev, et~al.
\newblock The kinetics human action video dataset.
\newblock {\em arXiv}, 2017.

\bibitem{kopuklu2019yowo}
Okan K{\"o}p{\"u}kl{\"u}, Xiangyu Wei, and Gerhard Rigoll.
\newblock You only watch once: A unified cnn architecture for real-time spatiotemporal action localization.
\newblock In {\em arXiv}, 2019.

\bibitem{Kuhn1955Hungarian}
Harold~W. Kuhn.
\newblock {The Hungarian Method for the Assignment Problem}.
\newblock {\em Naval Research Logistics Quarterly}, 2(1--2):83--97, March 1955.

\bibitem{li2020ava}
Ang Li, Meghana Thotakuri, David~A Ross, Jo{\~a}o Carreira, Alexander Vostrikov, and Andrew Zisserman.
\newblock The ava-kinetics localized human actions video dataset.
\newblock {\em arXiv}, 2020.

\bibitem{li2023unmasked}
Kunchang Li, Yali Wang, Yizhuo Li, Yi Wang, Yinan He, Limin Wang, and Yu Qiao.
\newblock Unmasked teacher: Towards training-efficient video foundation models.
\newblock In {\em ICCV}, 2023.

\bibitem{li2021improved}
Yanghao Li, Chao-Yuan Wu, Haoqi Fan, Karttikeya Mangalam, Bo Xiong, Jitendra Malik, and Christoph Feichtenhofer.
\newblock {MViTv2}: Improved multiscale vision transformers for classification and detection.
\newblock In {\em CVPR}, 2022.

\bibitem{cocodataset}
Tsung{-}Yi Lin, Michael Maire, Serge~J. Belongie, Lubomir~D. Bourdev, Ross~B. Girshick, James Hays, Pietro Perona, Deva Ramanan, Piotr Doll{'{a} }r, and C.~Lawrence Zitnick.
\newblock Microsoft {COCO:} common objects in context.
\newblock In {\em arXiv}, 2014.

\bibitem{lin2017feature}
Tsung-Yi Lin, Piotr Doll{\'a}r, Ross Girshick, Kaiming He, Bharath Hariharan, and Serge Belongie.
\newblock Feature pyramid networks for object detection.
\newblock In {\em CVPR}, 2017.

\bibitem{lin2014microsoft}
Tsung-Yi Lin, Michael Maire, Serge Belongie, James Hays, Pietro Perona, Deva Ramanan, Piotr Doll{\'a}r, and C~Lawrence Zitnick.
\newblock Microsoft coco: Common objects in context.
\newblock In {\em ECCV}, 2014.

\bibitem{loshchilov2017decoupled}
Ilya Loshchilov and Frank Hutter.
\newblock Decoupled weight decay regularization.
\newblock {\em arXiv}, 2017.

\bibitem{minderer2022simple}
Matthias Minderer, Alexey Gritsenko, Austin Stone, Maxim Neumann, Dirk Weissenborn, Alexey Dosovitskiy, Aravindh Mahendran, Anurag Arnab, Mostafa Dehghani, Zhuoran Shen, Xiao Wang, Xiaohua Zhai, Thomas Kipf, and Neil Houlsby.
\newblock Simple open-vocabulary object detection with vision transformers.
\newblock In {\em ECCV}, 2022.

\bibitem{pan2021actor}
Junting Pan, Siyu Chen, Mike~Zheng Shou, Yu Liu, Jing Shao, and Hongsheng Li.
\newblock Actor-context-actor relation network for spatio-temporal action localization.
\newblock In {\em CVPR}, 2021.

\bibitem{paszke2019pytorch}
Adam Paszke, Sam Gross, Francisco Massa, Adam Lerer, James Bradbury, Gregory Chanan, Trevor Killeen, Zeming Lin, Natalia Gimelshein, Luca Antiga, et~al.
\newblock Pytorch: An imperative style, high-performance deep learning library.
\newblock {\em NeurIPS}, 2019.

\bibitem{ren2015faster}
Shaoqing Ren, Kaiming He, Ross Girshick, and Jian Sun.
\newblock Faster r-cnn: Towards real-time object detection with region proposal networks.
\newblock In {\em NeurIPS}, 2015.

\bibitem{rezatofighi2019generalized}
Hamid Rezatofighi, Nathan Tsoi, JunYoung Gwak, Amir Sadeghian, Ian Reid, and Silvio Savarese.
\newblock Generalized intersection over union: A metric and a loss for bounding box regression.
\newblock In {\em CVPR}, 2019.

\bibitem{ryali2023hiera}
Chaitanya Ryali, Yuan-Ting Hu, Daniel Bolya, Chen Wei, Haoqi Fan, Po-Yao Huang, Vaibhav Aggarwal, Arkabandhu Chowdhury, Omid Poursaeed, Judy Hoffman, Jitendra Malik, Yanghao Li, and Christoph Feichtenhofer.
\newblock Hiera: A hierarchical vision transformer without the bells-and-whistles.
\newblock In {\em arXiv}, 2023.

\bibitem{shou2017}
Zheng Shou, Jonathan Chan, Alireza Zareian, Kazuyuki Miyazawa, and Shih-Fu Chang.
\newblock Cdc: Convolutional-de-convolutional networks for precise temporal action localization in untrimmed videos.
\newblock In {\em CVPR}, 2017.

\bibitem{shou2016}
Zheng Shou, Dongang Wang, and Shih-Fu Chang.
\newblock Temporal action localization in untrimmed videos via multi-stage cnns.
\newblock In {\em CVPR}, 2016.

\bibitem{singh2017online}
Gurkirt Singh, Suman Saha, Michael Sapienza, Philip~HS Torr, and Fabio Cuzzolin.
\newblock Online real-time multiple spatiotemporal action localisation and prediction.
\newblock In {\em ICCV}, 2017.

\bibitem{soomro2012ucf101}
Khurram Soomro, Amir~Roshan Zamir, and Mubarak Shah.
\newblock Ucf101: A dataset of 101 human actions classes from videos in the wild.
\newblock {\em arXiv}, 2012.

\bibitem{stewart2016end}
Russell Stewart, Mykhaylo Andriluka, and Andrew~Y Ng.
\newblock End-to-end people detection in crowded scenes.
\newblock In {\em CVPR}, 2016.

\bibitem{Sui2022ASA}
Lin Sui, Chen-Lin Zhang, Lixin Gu, and Feng Han.
\newblock A simple and efficient pipeline to build an end-to-end spatial-temporal action detector.
\newblock In {\em WACV}, 2022.

\bibitem{sun2018actor}
Chen Sun, Abhinav Shrivastava, Carl Vondrick, Kevin Murphy, Rahul Sukthankar, and Cordelia Schmid.
\newblock Actor-centric relation network.
\newblock In {\em ECCV}, 2018.

\bibitem{tang2020asynchronous}
Jiajun Tang, Jin Xia, Xinzhi Mu, Bo Pang, and Cewu Lu.
\newblock Asynchronous interaction aggregation for action detection.
\newblock In {\em ECCV}, 2020.

\bibitem{tong2022videomae}
Zhan Tong, Yibing Song, Jue Wang, and Limin Wang.
\newblock Video{MAE}: Masked autoencoders are data-efficient learners for self-supervised video pre-training.
\newblock In {\em NeurIPS}, 2022.

\bibitem{ulutan2020actor}
Oytun Ulutan, Swati Rallapalli, Carlos Torres, Mudhakar Srivatsa, and B Manjunath.
\newblock Actor conditioned attention maps for video action detection.
\newblock In {\em WACV}, 2020.

\bibitem{wang2023videomae}
Limin Wang, Bingkun Huang, Zhiyu Zhao, Zhan Tong, Yinan He, Yi Wang, Yali Wang, and Yu Qiao.
\newblock Videomae v2: Scaling video masked autoencoders with dual masking.
\newblock In {\em CVPR}, 2023.

\bibitem{wu2019long}
Chao-Yuan Wu, Christoph Feichtenhofer, Haoqi Fan, Kaiming He, Philipp Krahenbuhl, and Ross Girshick.
\newblock Long-term feature banks for detailed video understanding.
\newblock In {\em CVPR}, 2019.

\bibitem{wu2022}
Chao-Yuan Wu, Yanghao Li, Karttikeya Mangalam, Haoqi Fan, Bo Xiong, Jitendra Malik, and Christoph Feichtenhofer.
\newblock Memvit: Memory-augmented multiscale vision transformer for efficient long-term video recognition.
\newblock In {\em CVPR}, 2022.

\bibitem{wu2020context}
Jianchao Wu, Zhanghui Kuang, Limin Wang, Wayne Zhang, and Gangshan Wu.
\newblock Context-aware {RCNN}: A baseline for action detection in videos.
\newblock In {\em ECCV}, 2020.

\bibitem{wu2023stmixer}
Tao Wu, Mengqi Cao, Ziteng Gao, Gangshan Wu, and Limin Wang.
\newblock Stmixer: A one-stage sparse action detector.
\newblock In {\em CVPR}, 2023.

\bibitem{yang2017}
Ke Yang, Peng Qiao, Dongsheng Li, Shaohe Lv, and Yong Dou.
\newblock Exploring temporal preservation networks for precise temporal action localization.
\newblock In {\em AAAI}, 2017.

\bibitem{zhang2019structured}
Yubo Zhang, Pavel Tokmakov, Martial Hebert, and Cordelia Schmid.
\newblock A structured model for action detection.
\newblock In {\em CVPR}, 2019.

\bibitem{zhao2021tuber}
Jiaojiao Zhao, Xinyu Li, Chunhui Liu, Shuai Bing, Hao Chen, Cees~GM Snoek, and Joseph Tighe.
\newblock Tuber: Tube-transformer for action detection.
\newblock {\em arXiv}, 2021.

\end{thebibliography}
